\documentclass[preprint,12pt]{elsarticle}

\usepackage{amssymb}
\usepackage{booktabs}
\usepackage{amsmath}
\usepackage{multirow}
\usepackage{algorithm} 
\usepackage{algorithmic} 
\usepackage[normalem]{ulem} 
\useunder{\uline}{\ul}{} 
\usepackage{natbib} 

\journal{Neurocomputing}

\begin{document}

\begin{frontmatter}



\title{CRTrack: Low-Light Semi-Supervised Multi-object Tracking Based on Consistency Regularization}


\author[1]{Zijing Zhao}
\ead{21112202@bjtu.edu.cn}
\author[1]{Jianlong Yu}
\ead{24121438@bjtu.edu.cn}
\author[2]{Lin Zhang}
\ead{zhanglin20@sdnu.edu.cn}
\author[1]{Shunli Zhang\corref{cor1}} 
\ead{slzhang@bjtu.edu.cn} 
\cortext[cor1]{Corresponding author}

\affiliation[1]{organization={School of Software Technology, Beijing Jiaotong University},
            addressline={No.3 Shangyuan Village, Haidian District}, 
            city={Beijing},
            postcode={100044}, 
            state={Beijing},
            country={China}}

\affiliation[2]{organization={School of Information Science and Engineering, Shandong Normal University},
            addressline={No.88 Wenhua East Road, Lixia District}, 
            city={Jinan},
            postcode={250014}, 
            state={Shandong},
            country={China}}

\begin{abstract}
Multi-object tracking under low-light environments is prevalent in real life. Recent years  have seen rapid development in the field of multi-object tracking. However, due to the lack of datasets and the high cost of annotations, multi-object tracking under low-light environments remains a persistent challenge. In this paper, we focus on multi-object tracking under low-light conditions. To address the issues of limited data and the lack of dataset, we first constructed a low-light multi-object tracking dataset (LLMOT). This dataset comprises data from MOT17 that has been enhanced for nighttime conditions as well as multiple unannotated low-light videos. Subsequently, to tackle the high annotation costs and address the issue of image quality degradation, we propose a semi-supervised multi-object tracking method based on consistency regularization named CRTrack. First, we calibrate a consistent adaptive sampling assignment to replace the static IoU-based strategy, enabling the semi-supervised tracking method to resist noisy pseudo-bounding boxes. Then, we design a adaptive semi-supervised network update method, which effectively leverages unannotated data to enhance model performance. Dataset and Code: https://github.com/ZJZhao123/CRTrack.
\end{abstract}

\begin{highlights}
\item We create the first low-light multi-object tracking dataset (LLMOT) specifically for pedestrians.
\item We propose CRTrack, a low-light multi-object tracking method based on "teacher-student" network, which can leverage both labeled and unannotated data for training in a semi-supervised manner.
\end{highlights}


\end{frontmatter}


\section{Introduction}
Multi-object tracking (MOT) is a fundamental mid-level task in computer vision. By analyzing video sequences, MOT aims to localize multiple objects and associate them across frames to produce individual trajectories \cite{li2025unisort,zhang2025atptrack}. It is widely applied in domains such as action recognition \cite{lin2020gait,lin2021gait}, video analytics \cite{choi2012unified}, elderly care, and human-computer interaction \cite{geiger2012we}. Recently, practical MOT use cases have gained attention, further driving MOT development. However, most existing methods are designed for high-quality input data, overlooking common low-light conditions in real-world scenarios.

\begin{figure}[!t]
\centering
\includegraphics[width=3.2in]{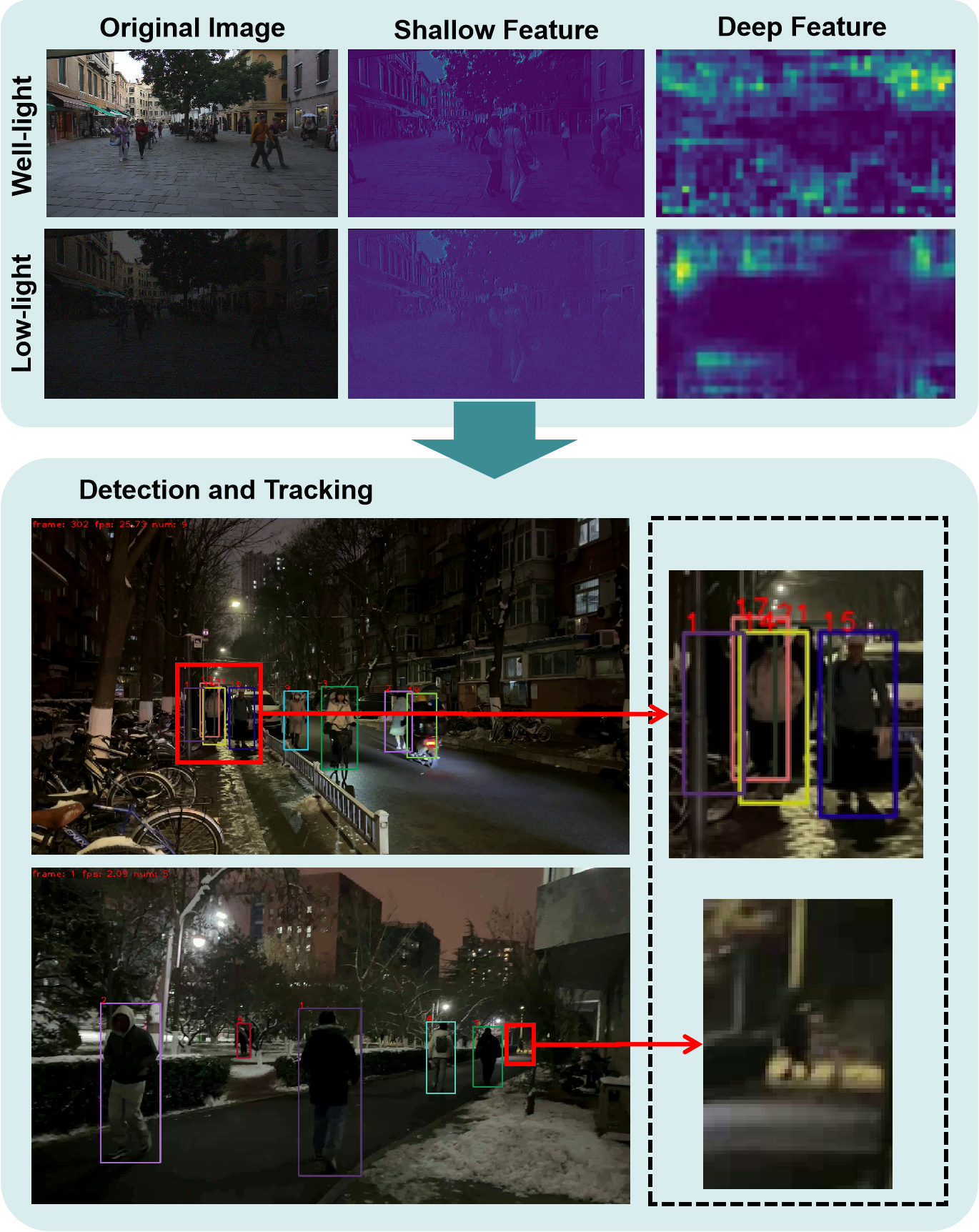}
\caption{Some results of low-light tracking. Comparing feature maps of normal and low-light images shows that there are fewer features in low-light environments. This results in poor performance. Tracking models trained on normal lighting datasets perform poorly when directly applied to low-light settings, which leads to missed and false detections.}
\label{fig_11}
\end{figure}

MOT in low-light environments is crucial for numerous applications. However, the dependence on annotated data by supervised methods is impractical, as low-light videos with corresponding annotations are scarce. Due to increased noise in low-light data compared to normal lighting, features become significantly harder to extract. As a result, tracking algorithms trained on normal-light scenes perform poorly in low-light environments, suffering from missed detections and false detections, as shown in Fig. \ref{fig_11}.

Due to the physical limitations of existing cameras, the video quality captured in low-light environments is inferior \cite{wang2024multi,yi2024comprehensive}. To avoid motion blur, cameras must shorten exposure times, limiting the amount of light collected. Meanwhile, increasing the image sensor's gain to enhance the signal introduces noise, further degrading image quality. Additionally, the low contrast in nighttime environments restricts the dynamic range and causes detail loss. These factors make it difficult for low-light videos to achieve the same level of clarity and detail as daytime recordings.

Low-light environments present two key challenges for multi-object tracking (MOT). First, low brightness and high noise in such conditions blur object boundaries, hindering the collection and annotation of tracking datasets. Second, low-light MOT itself faces numerous technical difficulties. Popular MOT methods are often detection-based, where objects are first detected and then associated across frames \cite{bewley2016simple,wojke2017simple,wang2020towards,zhang2021fairmot,gao2023dettrack,zhang2022bytetrack,cao2023observation,maggiolino2023deep}. These methods typically require high-quality annotated images as input but perform poorly on low-light images. Furthermore, they rely on supervised learning, making them unsuitable for unlabeled data. Some approaches apply pre-trained tracking algorithms directly to low-light scenarios, but they often fail to deliver satisfactory results. Other methods enhance visual quality by adding low-light restoration modules \cite{jin2024effective,wang2024multi,yi2024comprehensive,jones2023semi,yang2023track,teichman2012tracking}, while incurring additional computational costs and lacking robustness in complex low-light environments.

In this context, semi-supervised algorithms combine small amounts of labeled data with large amounts of unlabeled data, alleviating annotation scarcity and improving model generalization in low-light environments. Compared to fully supervised methods, semi-supervised approaches exploit the latent information in unlabeled data, offering a solution for low-light MOT. Although semi-supervised methods are widely applied in many fields \cite{chen2022label,chen2021temporal,zhou2021instant}, their application to low-light MOT is limited. This makes introducing semi-supervised methods into tracking tasks a valuable direction for further exploration.

In this paper, we present the Low-Light Multi-Object Tracking dataset (LLMOT), which focuses on pedestrian tracking and is specifically designed to address the challenges of tracking in low-light scenarios. We collected 6,264 unannotated frames through web scraping and real-world video recordings and developed a data augmentation method to adapt the MOT17\cite{milan2016mot16} dataset to nighttime conditions, thereby generating a fully annotated dataset for both training and testing purposes.

Based on the LLMOT dataset, we propose a low-light multi-object tracking method called CRTrack, which is built upon a consistency regularization approach. Existing methods are constrained by the limited availability of labeled data and face difficulties in effectively utilizing unlabeled data, particularly in low-light environments where relevant datasets are scarce. To address these limitations, we introduce a "teacher-student" semi-supervised learning framework that leverages consistency regularization to utilize both labeled and unlabeled data, achieving high-quality object detection and accurate tracking. To mitigate the issue of noisy pseudo-bounding boxes generated by the teacher network, we design a consistent adaptive sampling assignment strategy. Furthermore, we propose an adaptive semi-supervised network update mechanism to further enhance the training process, making the framework more robust and effective in handling low-light scenarios.

In summary, our main contributions are as follows:
\begin{itemize}
\item We present the first low-light multi-object tracking dataset specifically for pedestrians, which includes a large number of unannotated nighttime pedestrian videos and labeled nighttime data obtained by enhancing the MOT17\cite{milan2016mot16} dataset.
\item We propose CRTrack, a low-light multi-object tracking method based on a "teacher-student" network framework, which is capable of leveraging both labeled and unannotated data for semi-supervised training in low-light scenarios.
\item We conduct a comprehensive analysis of the proposed dataset and method. Experimental results demonstrate the effectiveness of our method, highlighting its superior performance and competitiveness in tracking tasks under low-light conditions.
\end{itemize}

\section{Related Work}
\subsection{Consistency Regularization}
Most object detection methods rely on large-scale and well-annotated datasets to achieve cutting-edge performance \cite{russakovsky2015imagenet,lin2014microsoft}. However, the high cost of acquiring labeled data drives the development of semi-supervised learning (SSL) methods, which leverage unlabeled data to significantly improve model performance \cite{berthelot2019remixmatch,sohn2020fixmatch,xie2020unsupervised,pham2021meta}.

Consistency regularization is among the most widely used techniques in SSL and requires models to produce consistent predictions across augmented versions of the same input \cite{bachman2014learning,sajjadi2016regularization}. Expanding on this concept, numerous studies optimize model performance through pseudo-label generation and distribution alignment.

For example, FixMatch \cite{sohn2020fixmatch} introduces a high-confidence pseudo-labeling strategy, retaining only pseudo-labels with sufficient confidence to improve efficiency. FlexMatch \cite{zhang2021flexmatch} dynamically adjusts thresholds for different classes to accommodate varying learning difficulties. LaplaceNet \cite{sellars2022laplacenet} leverages graph structures to optimize pseudo-label generation, while DC-SSL \cite{zhao2022dc} aligns predicted distributions with true distributions to improve pseudo-label accuracy. MW-FixMatch \cite{zheng2025mw} addresses class imbalance in SSL by incorporating a weight network to balance contributions from labeled and unlabeled data, which improves performance on imbalanced datasets.

Although these methods enhance model performance on unlabeled data, their effectiveness in specific scenarios, such as low-light environments, remains underexplored. Furthermore, pseudo-label quality is constrained by issues like confirmation bias, which may lead models to overfit on incorrect pseudo-labels \cite{wang2023conflict}.

To address these challenges, this study proposes a semi-supervised learning method, specifically designed for low-light pedestrian tracking, based on consistency regularization. Nighttime data augmentation strategies generate effective perturbations to enhance the model’s adaptability to low-light conditions. By jointly leveraging labeled and unlabeled data, the proposed method enhances both the robustness and accuracy of object detection and tracking in low-light scenarios. Compared to existing methods, this work introduces a specialized data augmentation strategy tailored for low-light conditions. It also improves pseudo-label generation and utilization within the consistency regularization framework, effectively overcoming the challenges of leveraging unlabeled data in such environments.

\subsection{Multi Object Tracking}
Current multi-object tracking algorithms primarily focus on improving the algorithms, paying little attention to data-related issues and diverse application scenarios. Detection-based tracking (DBT) is one of the most widely studied approaches for pedestrian tracking, detecting objects and performing data association across frames. DBT methods often combine motion and appearance features to improve tracking robustness.

Motion-based association methods predict object trajectories over time to match detections across frames. For example, the classic SORT\cite{bewley2016simple} algorithm uses Kalman filtering and IoU for target matching. Subsequent methods like BoT-SORT\cite{aharon2022bot} improve robustness to camera motion, while ByteTrack\cite{zhang2022bytetrack} enhances low-confidence target association.
Appearance-based association methods use visual features such as color, texture, and shape to match and track objects. However, they are less common due to their sensitivity to lighting and pose changes. QDTrack\cite{fischer2023qdtrack} employs contrastive learning to improve object association accuracy without relying on motion priors.
Combined motion-appearance methods integrate motion and appearance cues, yielding better performance in handling occlusions and complex motion patterns\cite{rajapaksha2024consistency}. For example, DeepSORT\cite{wojke2017simple} combines appearance features with motion cues, while FairMOT\cite{zhang2021fairmot} unifies detection and re-identification tasks to improve both accuracy and real-time efficiency. DetTrack\cite{gao2023dettrack} uses spatiotemporal features to mitigate the negative effects of occlusion. GeneralTrack\cite{qin2024towards} introduces a "point-wise to instance-wise relation" framework, enhancing generalizability across diverse MOT scenarios by improving object association robustness.
Additionally, unified cue integration methods\cite{li2025unisort} enhance performance in crowded and occlusion-heavy scenarios. These approaches leverage weak cues (e.g., confidence, pseudo-depth, and height) to compensate for failures of strong cues and refine detection associations, improving overall robustness.

Despite significant algorithmic advancements, DBT methods remain constrained by the performance of the detector, which is inherently constrained by the quality and scale of annotated datasets. Collecting and annotating pedestrian tracking datasets is particularly challenging due to large deformations, frequent occlusions, and numerous small objects. For nighttime pedestrian datasets, noise and blur significantly complicate data collection and reduce data quality.

To mitigate the challenges posed by limited annotated data, semi-supervised learning (SSL) methods are extensively studied in recent years. These methods combine labeled and unlabeled data to enhance performance while reducing the reliance on fully labeled datasets. For instance, some approaches use tracking information to iteratively train classifiers and extract useful samples from unlabeled data. These methods achieve accuracy comparable to fully supervised approaches while requiring significantly fewer labeled samples\cite{teichman2012tracking}. Additionally, multi-classifier systems that integrate supervised and semi-supervised models prove effective in enhancing tracking robustness under challenging scenarios, such as occlusions or visually similar objects. These systems also help reduce drifting errors\cite{stalder2009beyond}.

SSL methods also leverage temporal and spatial features to improve appearance model learning. For example, contrastive loss-based embedding methods learn from partially labeled videos, producing discriminative appearance features that enhance tracking robustness\cite{li2021semi}. Similarly, super-trajectory-based video segmentation approaches group trajectories with consistent motion patterns. These methods enable reliable propagation of initial annotations and improve robustness in handling occlusions\cite{wang2018semi}. However, these methods often rely on trajectory consistency, but this consistency is often unreliable in low-light or noisy environments, limiting their adaptability.

Existing nighttime multi-object tracking methods\cite{yi2024comprehensive,wang2024multi} often use data augmentation on limited datasets to simulate low-light conditions, but they lack real-world scene data and are constrained by dataset size. To address these issues, we propose a method for pedestrian tracking in low-light conditions that leverages annotated and unannotated data during training. Drawing inspiration from SSL methods such as iterative training strategies\cite{teichman2012tracking} and contrastive learning mechanisms\cite{li2021semi}, our approach addresses their limitations in noisy environments. It also improves adaptability to real-world scenarios, overcoming the constraints imposed by dataset size.

\section{Dataset Construction}
To advance multi-object tracking in low-light conditions, we developed a low-light pedestrian tracking dataset, LLMOT. The dataset includes a total of 11,580 images, comprising 5,316 labeled images and 6,264 unlabeled images.

The labeled data is derived from the MOT17 dataset, which we enhanced to simulate nighttime conditions. The MOT17 dataset contains 14 sequences, with 7 sequences annotated for the pedestrian class. It covers various scenes and environments, captured from different perspectives, under diverse lighting conditions, and with varying crowd densities. We selected the 7 labeled video sequences from MOT17 for augmentation in our dataset. These sequences are evenly divided into two subsets: one for labeled training in semi-supervised learning and the other for final validation.

To ensure realistic enhancement and preserve original object details, we evaluated three methods: stable-diffusion\cite{rombach2022high}, CycleGAN\cite{almahairi2018augmented}, and a traditional manually designed approach. Based on our comparison, we find that the manually designed approach provides the most realistic results and best preserves the original pedestrian information in the images. Consequently, we selected this method for feature enhancement.

\begin{figure}[!t]
\centering
\includegraphics[width=3.2in]{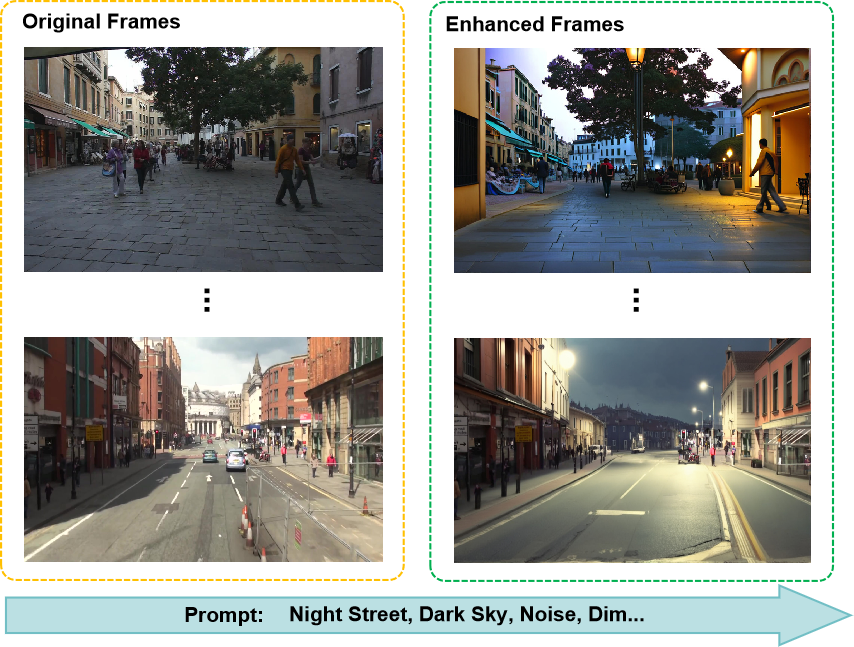}
\caption{The results of enhancement using the stable-diffusion model. The image on the left is the original, and the results on the right is generated by guiding the stable-diffusion model with prompts like night street, dark sky, noise, dim...}
\label{fig_1}
\end{figure}

When using the stable-diffusion model, while it achieves style transformation, it results in significant detail loss, causing small objects to disappear and rendering the original ground truth invalid. Some of the results are shown in Fig. \ref{fig_1}. Moreover, generating enhanced images with the stable-diffusion model is highly time-consuming, making it impractical for large-scale image generation tasks.

\begin{figure}[!t]
\centering
\includegraphics[width=2.1in]{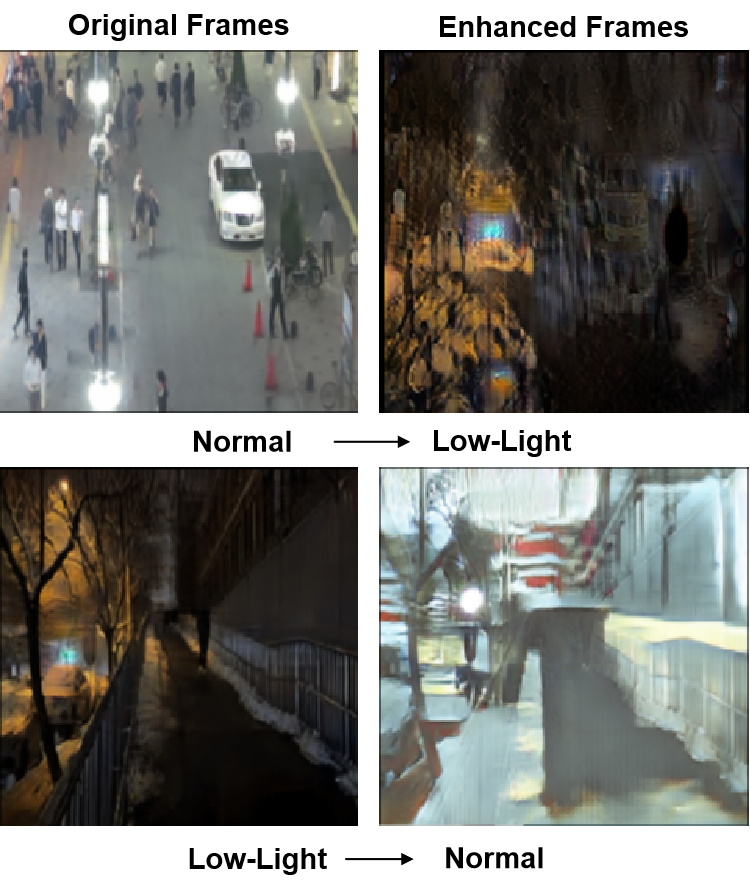}
\caption{The results of enhancement using CycleGAN. Through adversarial learning, low-light images are transformed into normal-light images, and normal-light images are transformed into low-light images.}
\label{fig_2}
\end{figure}

Using CycleGAN for processing may introduce artifacts during enhancement, as shown in Fig. \ref{fig_2}. The tracking dataset consists of multiple video sequences, and the repetition of fixed scenes within video frames often generates artifacts during enhancement. With CycleGAN, using a few training iterations results in insufficient detail learning and unrealistic scenes, while excessive iterations lead to severe artifact issues.

\begin{figure}[!t]
\centering
\includegraphics[width=2.7in]{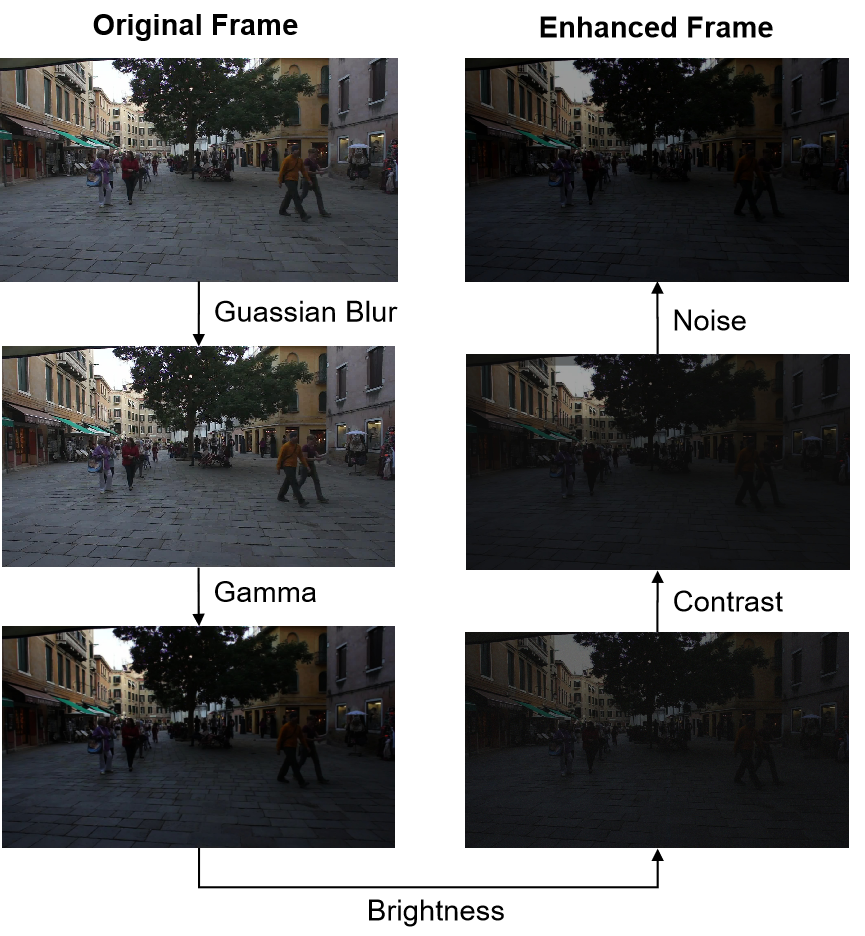}
\caption{The results of enhancement using manually designed method. We combine multiple enhancement methods, including Gaussian blur and gamma correction, and ultimately generated the low-light enhancement result.}
\label{fig_3}
\end{figure}

The manually designed enhancement method using traditional techniques produces the most natural results among the three methods. These traditional methods simulate various noises and low-light conditions by combining adjustments such as contrast modification, brightness reduction, Gaussian blur, gamma correction, and noise addition. As shown in Fig. \ref{fig_3}, we sequentially applied these adjustments to generate enhanced images. To introduce greater diversity in lighting and noise conditions, we incorporated randomization into the brightness adjustment, Gaussian blur, and noise addition processes.

The unannotated dataset consists of 10 video segments sourced from both web content and real-world scenarios, totaling 6,264 frames. It includes various low-light conditions, such as cloudy and nighttime scenes, spanning environments like regular roads, footbridges, and areas with significant railing obstructions.

\section{Methodology}
In this section, we introduce the overall framework of our low-light semi-supervised multi-object tracking method, CRTrack, which is built on consistency regularization. The proposed tracking framework consists of two main components: detection, which identifies objects in frames, and association, which links these objects across frames. First, we train a semi-supervised detector using a teacher-student network structure, which forms the backbone of the detection process by leveraging both labeled and unlabeled data. The semi-supervised detector includes two key modules: consistent adaptive sampling assignment and an adaptive update method. These modules are specifically designed to enhance the detector's performance in low-light conditions. Once trained, the detector is used to accurately identify objects in video frames, serving as the foundation for subsequent tracking. Subsequently, we perform object association to link detected objects across frames, ensuring temporal consistency and robust tracking. The details of these modules and methods will be elaborated in the following sections.
\subsection{Semi-Supervised Method}
\begin{figure}[!t]
\centering
\includegraphics[width=5.3in]{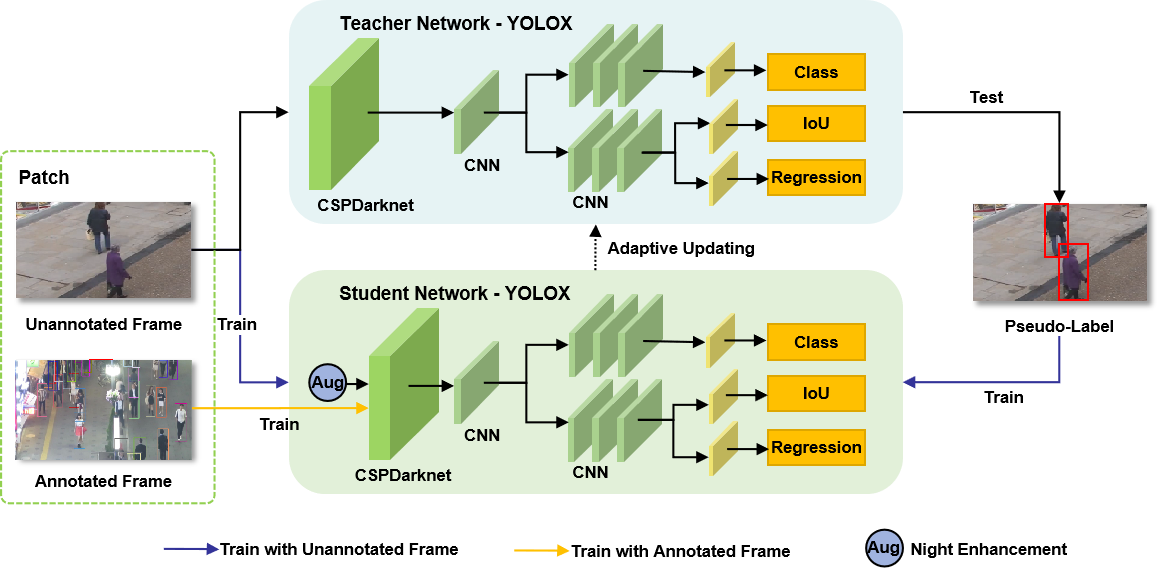}%
\hfil
\caption{Teacher-student network pipeline. Use the output of the teacher network as ground truth to guide the student network learning, and use the updates of the student network's parameters to fine-tune the teacher network's weights.}
\label{fig_4}
\end{figure}

In detection-based multi-object tracking frameworks, robust detection performance is crucial for ensuring accurate input data for subsequent data association. To address the challenge of limited labeled data and fully utilize pretrained models for improved training efficiency, we propose a semi-supervised learning method utilizing consistency regularization. We adopt the teacher-student network structure, a representative approach in semi-supervised learning, to enhance the original YOLOX\cite{ge2021yolox} object detection method, as illustrated in Fig. \ref{fig_4}. First, we construct a teacher-student network, where the teacher network performs detection and its outputs guide the training of the student network. The weights of the student network are then used to adaptively update and fine-tune the teacher network.

In our model, we use YOLOX object detection networks with identical pretrained parameters for both the teacher and student networks. A batch is then formed by combining annotated and unannotated data to train the student network.

For unannotated data, we pass the data through the teacher network for detection, selecting high-confidence predictions as pseudo-ground truth to train the student network. The student network's input is enhanced using a manually designed nighttime augmentation method. This augmentation strategy increases data diversity, enabling the student network to better capture data distribution and structure, improve model generalization, and enhance robustness against noise and interference. Consistency regularization is applied by augmenting the input data for the student network while keeping the teacher network's input data unchanged during training. Given an unannotated set $\mathcal{C}U = { x_i }{i=1}^M$ with $M$ frames, we maintain a teacher network $f_t(\cdot; \Theta_t)$ and a student network $f_s(\cdot; \Theta_s)$. The networks minimize the unannotated data training loss:
\begin{equation}
\label{deqn_ex1a}
\begin{split}
\mathcal{L}_U = \frac{1}{M}\sum_{i=1}^{M} \Big[ & \lambda_{cls}\mathcal{L}_{cls}(f_s(\mathcal{E}(x_i)),\hat{y}_i) \\
&+ \lambda_{reg}\mathcal{L}_{reg}(f_s(\mathcal{E}(x_i)),\hat{y}_i) \\
& + \lambda_{IoU}\mathcal{L}_{IoU}(f_s(\mathcal{E}(x_i)),\hat{y}_i) \Big].
\end{split}
\end{equation}
where $\mathcal{E}$ represents the low-light enhancement function, $\lambda$ is the weighting parameter, and $\hat{y}_i = f_t((x_i); \Theta_t)$ denotes the pseudo-bounding boxes (pseudo-bboxes) generated by the teacher network.

For annotated data, we directly input them into the student network for supervised training. The labeled data serves as an accurate baseline, enabling the model to better fit the data distribution. Given a labeled set $\mathcal{C}L = { x_j }{j=1}^N$ with $N$ frames, the labeled data training loss is minimized as:
\begin{equation}
\label{deqn_ex1a}
\begin{split}
\mathcal{L}_L = \frac{1}{N}\sum_{j=1}^{N} \Big[ & \lambda_{cls}\mathcal{L}_{cls}(f_s(x_j),y_j) + \lambda_{reg}\mathcal{L}_{reg}(f_s((x_j),y_j) \\
& + \lambda_{IoU}\mathcal{L}_{IoU}(f_s(x_j),y_j) \Big].
\end{split}
\end{equation}
where $y_j$ represents the ground truth (GT) of the $j$-th image.

The total loss is calculated as:
\begin{equation}
\label{deqn_ex1a}
\mathcal{L}_T = \lambda_{U}\mathcal{L}_U + \mathcal{L}_L.
\end{equation}
where $\lambda_{U}$ represents the weight parameter for the unannotated loss, determined by the proportions of annotated and unannotated data in the batch.

\subsection{Consistent Adaptive Sampling Assignment}
In the teacher network, only targets with detection and classification confidence exceeding a certain threshold are used as ground truth (GT) for the student network. However, this static label assignment violates a crucial property of semi-supervised learning: pseudo-labels may sometimes contradict the network's own predictions. Given an object $u$ in frame $x$, denoted as $x^u$, we take regression as an example. The instance-level pseudo-label satisfies:
\begin{equation}
\label{deqn_ex1a}
\hat{g}=\underset{g}{\arg\min}(f_t((x^u),g)
\end{equation}
where $g$ represents a candidate pseudo-label, serving as a variable in the optimization process. The final pseudo-label $\hat{g}$ is obtained by minimizing the loss function $f_t$, ensuring that $\hat{g}$ is the pseudo-label that best fits the teacher network's prediction. Ideally, $\hat{g}$ and $g$ should be as close as possible. However, when adopting a static pseudo-label generation method, this rule can be violated, leading to inconsistencies between pseudo-labels and predictions.

To address this issue, we propose assigning pseudo-labels that minimize their corresponding loss:
\begin{equation}
\label{deqn_ex1a}
\begin{split}
{\min}\sum_{i=1}^{M} \Big[ & \lambda_{cls}\mathcal{L}_{cls}(f_s(\mathcal{E}(x_i^u)),\hat{y}_i^u) \\
&+ \lambda_{reg}\mathcal{L}_{reg}(f_s(\mathcal{E}(x_i^u)),\hat{y}_i^u)\\
& + \lambda_{IoU}\mathcal{L}_{IoU}(f_s(\mathcal{E}(x_i^u)),\hat{y}_i^u) \Big].
\end{split}
\end{equation}
where $\mathcal{E}$ represents the low-light enhancement function.

We assign predictions with the lowest losses as positive pseudo-labels, aiming to balance the trade-off between the quantity and quality of pseudo-labels. Although this may reduce the number of pseudo-labels, their quality is significantly improved, effectively reducing noise interference and enhancing both training efficiency and model performance. Additionally, predictions with losses exceeding a certain threshold are assigned as negative samples, preventing uncertain samples from interfering with training. This approach clarifies the optimization objective, provides high-quality supervisory signals, minimizes the impact of low-quality pseudo-labels, and ensures efficient and reliable training.
YOLOX adopts an anchor-free design to directly predict the center points and sizes of objects. However, it does not fully utilize center point information during the loss calculation, which limits the effectiveness of matching. To address this, we leverage center point information to calculate the matching cost between each prediction and pseudo-label. Predictions with the lowest costs are considered positive. Given a prediction $p_n$, the matching cost $\mathcal{C}_{nk}$ between $p_n$ and a pseudo-bounding box $y_k$ is calculated as:
\begin{equation}
\label{deqn_ex1a}
\begin{split}
\mathcal{C}_{nk} = \Big[ & \lambda_{cls}\mathcal{L}_{cls}(p_n,y_k) + \lambda_{reg}\mathcal{L}_{reg}(p_n,y_k) \\
& + \lambda_{IoU}\mathcal{L}_{IoU}(p_n,y_k)+\lambda_{dis}\mathcal{C}_{dis} \Big].
\end{split}
\end{equation}
where $\lambda_{dis}$ is a weighting parameter, and $\mathcal{C}_{dis}$ calculates the distance between the center of prediction $p_n$ and pseudo-label $y_k$, serving as a center prior with a small weighting value to stabilize training.

Based on each pseudo-label's matching cost, predictions with the top $K$ lowest costs are assigned as positive. By aligning assignment with the model’s detection quality, this approach minimizes the impact of noise in pseudo-label assignments. This module is inspired by Consistent-Teacher\cite{wang2023consistent}, which employs an anchor-based structure. In contrast, our method adopts the anchor-free YOLOX framework, effectively alleviating detection shifts caused by noise in low-light environments and enhancing consistency between the student and teacher networks.

\subsection{Adaptive Network Updating}
In detection-based tracking methods, the detector is typically pretrained on a detection dataset and subsequently fine-tuned on a tracking dataset, resulting in a slow update process. In conventional detection and classification tasks, teacher networks are commonly updated using the exponential moving average (EMA) method during fine-tuning. However, in multi-object tracking, the student network is already fine-tuned. If the teacher network is updated solely using EMA, the update pace may be excessively slow, potentially causing the faster-updating student network to outperform the teacher network. As a result, the underperforming teacher network may provide low-quality pseudo-ground truth, reducing the student network's training effectiveness and triggering a feedback loop that degrades overall network performance. To address this issue, we propose an adaptive update method to dynamically update the teacher network.
\begin{algorithm}[H]
\caption{Adaptive Update Method.}\label{alg:alg1}
\begin{algorithmic}
\STATE bestModel = teacher
\STATE bestEval = eval(bestModel)
\WHILE{training}
    \STATE Train student for one epoch
    \STATE teacher = EMA(student)
    
    \STATE teacherEval = eval(teacher)
    \STATE studentEval = eval(student)
    
    \IF{teacherEval \textgreater\ bestEval}
        \STATE bestModel = teacher
        \STATE bestEval = teacherEval
    \ENDIF
    
    \IF{studentEval \textgreater\ bestEval}
        \STATE bestModel = student
        \STATE bestEval = studentEval
        \STATE teacher = student
    \ENDIF
\ENDWHILE
\end{algorithmic}
\label{alg1}
\end{algorithm}
As shown in Algorithm \ref{alg:alg1}, we evaluate the trained student network, the EMA-updated teacher network, and the previous best model using validation results. Based on the evaluation, we update the best-performing network and its corresponding validation score. The best network is then assigned to the teacher network to ensure optimal performance.
\subsection{Association}
In multi-object tracking (MOT), incorporating appearance features for data association enhances robustness and accuracy, especially when detection performance is suboptimal. Features such as color and texture help reduce ID switches, complement detection information, and improve tracking continuity and multi-target differentiation, ensuring stable tracking in complex environments.

In low-light environments, object features are diminished, boundaries are blurred, and environmental noise is high, resulting in poorer detection performance compared to normal lighting conditions. Relying solely on motion cues for data association in such conditions can lead to frequent ID switches. To address this, we propose a method based on OC-SORT\cite{cao2023observation} that integrates both appearance and motion features for data association.

\begin{figure}[!t]
\centering
\includegraphics[width=3.3in]{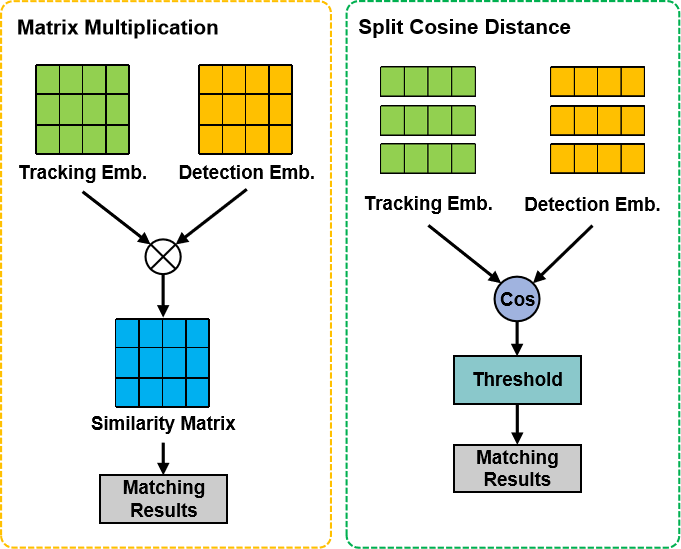}
\caption{Comparison between matrix multiplication and split cosine distance.}
\label{fig_13}
\end{figure}

Our method draws inspiration from Deep OC-SORT\cite{maggiolino2023deep} but employs a different approach for appearance feature association, as illustrated in Fig. \ref{fig_13}. In Deep OC-SORT, similarity between detection and tracking embeddings is computed directly through matrix multiplication for data association. While this method is efficient and suitable for large-scale data, it fails in low-light conditions. Blurred features and significant environmental noise in such scenarios can lead to misjudgments, making it difficult to accurately distinguish targets.
To address this issue, we adopt split cosine distance for similarity calculations. Specifically, we compute the cosine distance between detection and tracking embeddings in a pairwise manner, convert it into similarity scores, and identify the maximum similarity for each pair. A threshold is then applied to filter out unreliable matches, ensuring that only reliable similarities are retained. This approach offers significant advantages in low-light conditions, as it reduces the influence of feature magnitude variations and noise interference, focusing instead on the directional consistency of feature vectors to enhance robustness. Fig. \ref{fig_13} illustrates the comparison between matrix multiplication and split cosine distance.

\begin{figure}[!t]
\centering
\includegraphics[width=5.3in]{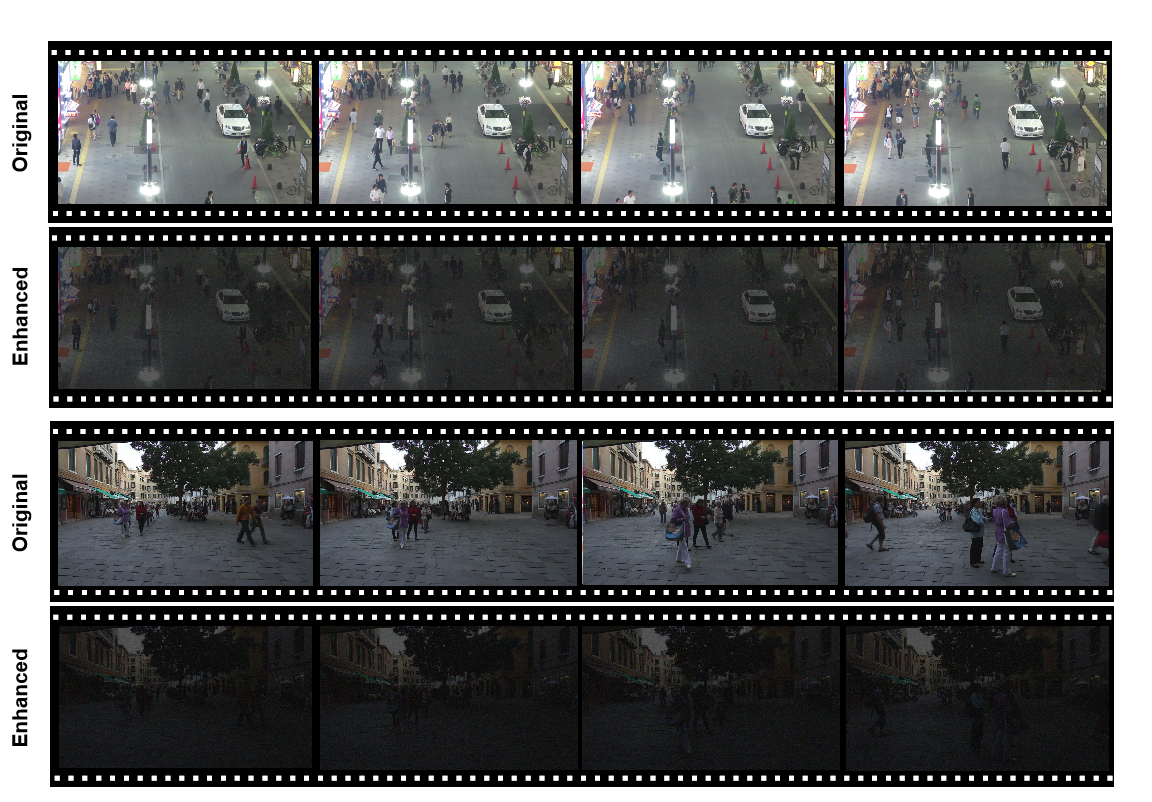}%
\hfil
\caption{Two example videos of enhanced MOT17 from our LLMOT dataset. The enhanced MOT17 has dimmer lighting and blurrier scenes, better matching the characteristics of low-light conditions, but the pedestrian objects remain in their original positions.}
\label{fig_5}
\end{figure}

\begin{figure}[!t]
\centering
\includegraphics[width=4.8in]{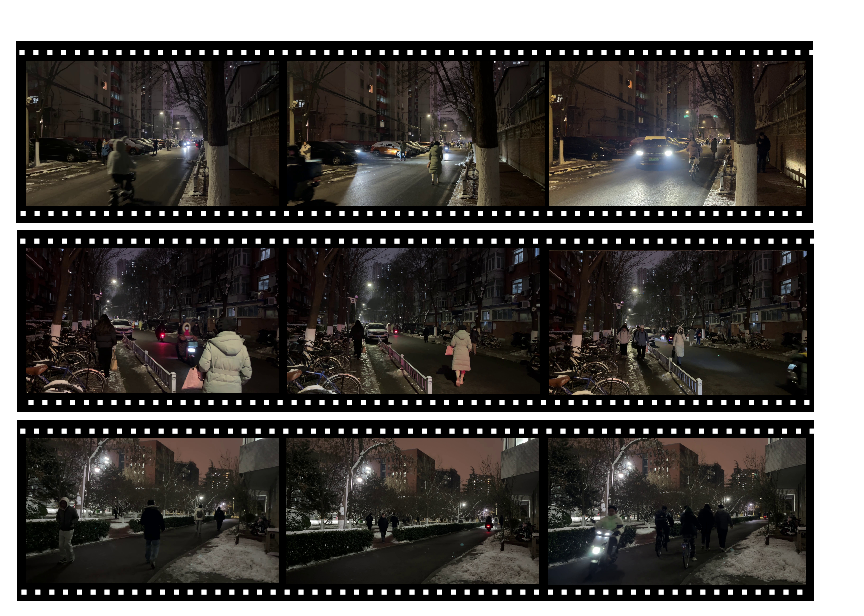}
\caption{Three example videos of unannotated data from our LLMOT dataset. Each row consists of three frames selected from a video.}
\label{fig_6}
\end{figure}

%
\begin{figure*}[!t]
\centering
\includegraphics[width=5.3in]{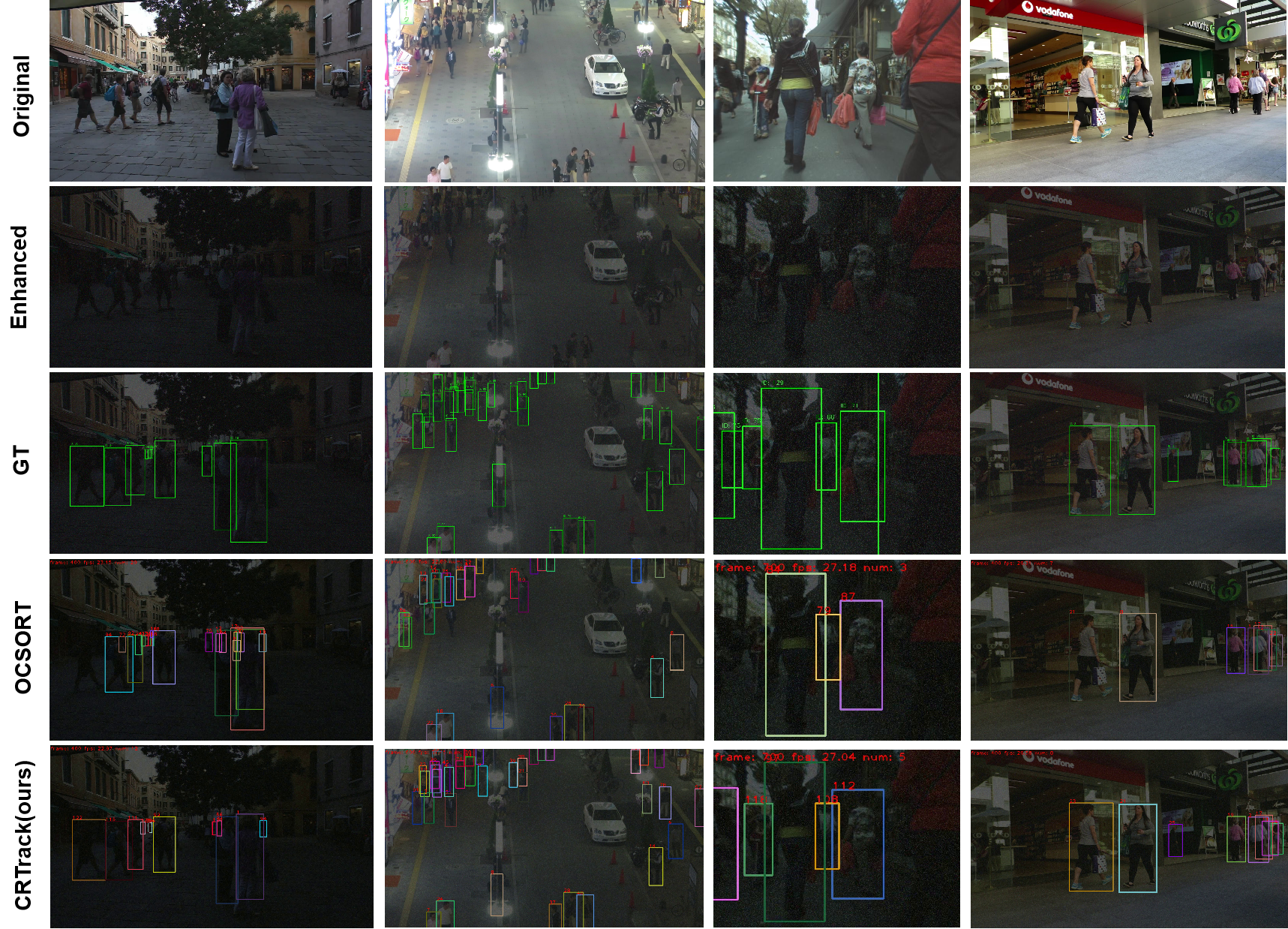}%
\hfil
\caption{Example results on the LLMOT validation set. Our CRTrack generates fewer tracking errors than OC-SORT\cite{cao2023observation}.}
\label{fig_7}
\end{figure*}

\section{Experiments}
\subsection{Experiment Setup}
{\bf{Dataset.}} Our experiments utilize two datasets: LLMOT, designed for low-light object tracking, and the CrowdHuman\cite{shao2018crowdhuman} dataset, used for joint training. LLMOT comprises MOT17 data enhanced to simulate low-light conditions and additional unannotated low-light data, as shown in Fig. \ref{fig_5} and Fig. \ref{fig_6}. CrowdHuman is a large-scale pedestrian detection dataset, containing 15,000 images in the training set, 5,000 images in the test set, and 4,370 images in the validation set. Incorporating CrowdHuman during training allows the model to learn diverse pedestrian features, thereby enhancing its robustness in low-light conditions.

{\bf{Evaluation Metrics.}} We evaluate object detection and tracking performance using the following metrics: detection accuracy (DetA)\cite{li2022learning}, which measures overall detection precision; multiple object tracking accuracy (MOTA)\cite{bernardin2008evaluating}, which accounts for false positives, missed detections, and ID errors; higher order tracking accuracy (HOTA)\cite{luiten10higher}, which provides a detailed analysis of detection and trajectory association; identity F1 score (IDF1)\cite{ristani2016performance}, which evaluates both detection accuracy and trajectory consistency; and association accuracy (AssA)\cite{li2022learning}, which focuses on target and trajectory association accuracy. Combined, these metrics comprehensively assess the tracking method's performance. Additionally, object detectors are evaluated using average precision (AP) and average recall (AR).

{\bf{Implementation Details.}} Following OC-SORT\cite{cao2023observation}, we adopt YOLOX\cite{ge2021yolox} pretrained on the COCO dataset as our detector and use the same association method as OC-SORT. OC-SORT emphasizes the role of observations in tracking, compensating for errors in trajectory extension caused by state estimation, making it a suitable choice for association in our method. Initially, we pre-train YOLOX using supervised learning on the CrowdHuman\cite{shao2018crowdhuman} dataset and the annotated training portion of our LLMOT dataset for 80 epochs. We then fine-tune the model using semi-supervised learning with the CrowdHuman dataset and the LLMOT training set for an additional 30 epochs. The semi-supervised training process takes approximately 18 hours on 4 Nvidia RTX 3090 GPUs, with 4 images processed per GPU (3 labeled and 1 unlabeled).
\subsection{Main Results}

\begin{table}
\begin{center}

\caption{Detection Performance Comparison Across Normal Lighting, Enhanced Low-light, and Mixed Low-light Data.}
\label{tab1}
\resizebox{\linewidth}{!}{
\begin{tabular}{c c c c c c c c c c c}
\toprule
Training Data & mAP & AP50 & AP75 & $\mathrm{AP}_{small}$  & $\mathrm{AP}_{medium}$  & $\mathrm{AP}_{large}$  & mAR & $\mathrm{AR}_{small}$& $\mathrm{AR}_{medium}$  & $\mathrm{AR}_{large}$\\
\midrule
NL & 46.5 & 73.6 & 52.3 & 7.1 & 36.5 & 60.3 & 50.7 & 8.4 & 40.9 & 65.0\\
LL & 52.8 & 79.1 & 59.8 & 12.7 & 41.9 & 66.8 & 57.9 & 18.3 & 48.8 & 70.7\\ 
LLMOT & \textbf{54.6} & \textbf{82.0} & \textbf{61.5} & \textbf{14.6} & \textbf{44.9} & \textbf{66.9} & \textbf{59.4} & \textbf{25.2} & \textbf{51.4} & \textbf{70.9}\\
\bottomrule
\end{tabular}}
\end{center}
\end{table}

\begin{figure}[!t]
\centering
\includegraphics[width=3.8in]{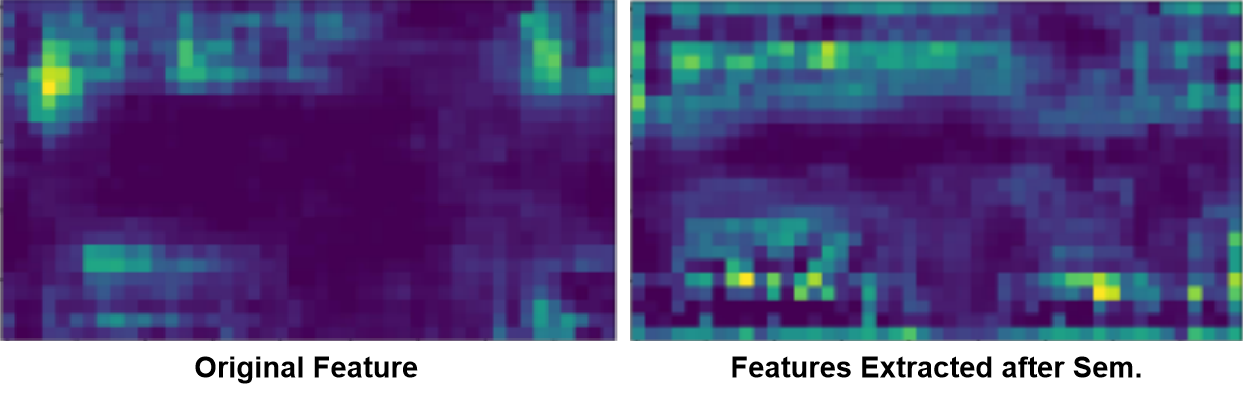}
\caption{Visualization of appearance features of LLMOT dataset. We can see that compared to models trained in normal-lighting dataset, our semi-supervised model(Sem.) can extract richer features from nighttime data.}
\label{fig_12}
\end{figure}

{\bf{Detection performance.}} We analyze the performance of detectors in low-light environments under three training settings: a detector trained on normal lighting data (NL), a detector trained on annotated low-light data (LL), and a detector trained on the LLMOT dataset using a semi-supervised algorithm. As shown in Fig. \ref{fig_12}, compared to models trained on normal lighting datasets, our model extracts richer features from nighttime data, enhancing detection performance under low-light conditions. The detection results are presented in Tab. \ref{tab1}. From Tab. \ref{tab1}, we observe that the detector trained on the LLMOT dataset using the semi-supervised algorithm achieves the best performance on low-light images. In contrast, the detector trained on normal lighting datasets shows poor robustness in low-light environments, making it difficult to transfer directly. Training on a limited annotated low-light dataset (the annotated portion of LLMOT) significantly improves detection performance. However, with the addition of semi-supervised learning and an increased amount of data, the detector achieves further improvements. These results demonstrate the effectiveness of our approach in enhancing detection performance under low-light conditions.


\begin{table}
\caption{The Performance of Tracking Methods in Low-light Environments. The upper section displays results from popular tracking algorithms using supervised learning, while the lower section presents outcomes from these algorithms and our approach under semi-supervised learning.}
\label{tab2}
\resizebox{\linewidth}{!}{
\begin{tabular}{ccccccccc}
\hline
Method                                                                  & Association       & Detection  & Training Data & DetA            & MOTA            & HOTA            & IDF1            & AssA            \\ \hline
\multirow{2}{*}{SORT~\cite{bewley2016simple}}                           & Motion            & YOLOX      & NL            & 44.534          & 52.594          & 51.233          & 63.293          & 59.208          \\
                                                                        & Motion            & YOLOX      & LL            & 55.798          & 64.550          & 58.059          & 69.771          & 60.908          \\
\multirow{2}{*}{OC-SORT~\cite{cao2023observation}}                      & Motion            & YOLOX      & NL            & 44.470          & 52.730          & 50.527          & 61.459          & 57.660          \\
                                                                        & Motion            & YOLOX      & LL            & 55.804          & 65.144          & 57.754          & 68.873          & 60.233          \\
\multirow{2}{*}{ByteTrack~\cite{zhang2022bytetrack}}                    & Motion            & YOLOX      & NL            & 50.742          & 61.215          & 55.638          & 68.493          & 61.317          \\
                                                                        & Motion            & YOLOX      & LL            & 59.561          & 70.52           & 60.381          & 72.847          & 61.616          \\
\multirow{2}{*}{DeepSORT~\cite{wojke2017simple}}                        & Motion+Appearance & YOLOX      & NL            & 48.694          & 58.547          & 53.677          & 66.257          & 59.460          \\
                                                                        & Motion+Appearance & YOLOX      & LL            & 58.263          & 68.124          & 58.85           & 70.031          & 60.059          \\
\multirow{2}{*}{MOTDT~\cite{chen2018real}}                              & Motion+Appearance & YOLOX      & NL            & 48.861          & 58.514          & 53.777          & 65.892          & 59.493          \\
                                                                        & Motion+Appearance & YOLOX      & LL            & 58.633          & 67.588          & 58.924          & 69.706          & 59.779          \\
\multirow{2}{*}{Deep OC-SORT~\cite{maggiolino2023deep}}                 & Motion+Appearance & YOLOX      & NL            & 47.886          & 57.152          & 53.581          & 65.694          & 60.224          \\
                                                                        & Motion+Appearance & YOLOX      & LL            & 58.272          & 69.282          & 60.608          & 74.265          & 63.335          \\
\multirow{2}{*}{GeneralTrack~\cite{qin2024towards}}                     & Motion+Appearance & YOLOX      & NL            & 50.838          & 61.313          & 55.635          & 68.452          & 61.671          \\
                                                                        & Motion+Appearance & YOLOX      & LL            & 59.729          & 70.196          & 60.381          & 72.959          & 61.876          \\ \hline
SORT~\cite{bewley2016simple}                                            & Motion            & Sem.(ours) & LLMOT         & 57.285          & 66.968          & 59.908          & 71.655          & 63.068          \\
OC-SORT~\cite{cao2023observation}                                       & Motion            & Sem.(ours) & LLMOT         & 57.190          & 66.883          & 59.978          & 72.171          & 63.256          \\
ByteTrack~\cite{zhang2022bytetrack}                                     & Motion            & Sem.(ours) & LLMOT         & \textbf{60.896} & \textbf{71.624} & 61.816          & 73.614          & 63.506          \\
DeepSORT~\cite{wojke2017simple}                                         & Motion+Appearance & Sem.(ours) & LLMOT         & 59.715          & 70.360          & 62.041          & 74.674          & 64.931          \\
MOTDT~\cite{chen2018real}                                               & Motion+Appearance & Sem.(ours) & LLMOT         & 60.016          & 70.583          & 61.346          & 74.039          & 63.183          \\
Deep OC-SORT~\cite{maggiolino2023deep}                                  & Motion+Appearance & Sem.(ours) & LLMOT         & 60.257          & 71.052          & 61.832          & 74.799          & 63.862          \\
GeneralTrack~\cite{qin2024towards}                                      & Motion+Appearance & Sem.(ours) & LLMOT         & 60.461          & 71.371          & {\ul 62.249}    & {\ul 74.982}    & {\ul 64.943}    \\
CRTrack (ours)                                                          & Motion+Appearance & Sem.(ours) & LLMOT         & {\ul 60.532}    & {\ul 71.544}    & \textbf{62.472} & \textbf{75.864} & \textbf{64.976} \\ \hline
\end{tabular}}
\end{table}

\begin{table}
\begin{center}
\caption{Ablation on Semi-supervised Learning (Sem.), Consistent Adaptive Sampling Assignment (ASA), Adaptive Network Update (ANU), Appearance (App.), and Split Cosine Distance (SCD).}
\label{tab3}
\resizebox{\linewidth}{!}{
\begin{tabular}{c c c c c c c c c c c}
\toprule
OCSORT & Sem. & ASA & ANU & App. & SCD & DetA & MOTA & HOTA & IDF1 & AssA \\
\midrule
\checkmark & & & & & & 55.804 & 65.144 & 57.754 & 68.873 & 60.233 \\
\checkmark & \checkmark & & & & & 55.729 & 65.923 & 58.428 & 70.29 & 61.585 \\ 
\checkmark & \checkmark & \checkmark & & & & 56.557 & 66.081 & 59.011 & 70.881 & 61.993 \\
\checkmark & \checkmark & \checkmark & \checkmark & & & 57.190 & 66.883 & 59.978 & 72.171 & 63.256 \\
\checkmark & \checkmark & \checkmark & \checkmark & \checkmark & & 60.257 & 71.052 & 61.832 & 74.799 & 63.862 \\
\checkmark & \checkmark & \checkmark & \checkmark & \checkmark & \checkmark & \textbf{60.532} & \textbf{71.544} & \textbf{62.472} & \textbf{75.864} & \textbf{64.976} \\
\bottomrule
\end{tabular}}
\end{center}
\end{table}

{\bf{Tracking performance.}} We analyze the tracking performance in low-light conditions. Several representative tracking algorithms are compared, including motion-based association algorithms such as SORT\cite{bewley2016simple}, OC-SORT\cite{cao2023observation}, and ByteTrack\cite{zhang2022bytetrack}, as well as algorithms that combine motion and appearance, such as DeepSORT\cite{wojke2017simple}, MOTDT\cite{chen2018real}, Deep OC-SORT\cite{maggiolino2023deep}, and GeneralTrack\cite{qin2024towards}. It is evident that the latter generally outperforms the former. In low-light conditions, detection performance is weaker compared to normal lighting. Relying solely on motion cues for data association often leads to ID switches, especially in cases of pedestrian occlusions. Integrating motion and appearance cues helps mitigate this issue.

We compared the tracking results of each algorithm under low-light conditions using detectors trained under three configurations: supervised training in normal lighting, supervised training in low-light, and semi-supervised training. As shown in Tab. \ref{tab1}, detectors trained in normal lighting struggle to adapt to low-light environments for tracking, underscoring the critical importance of training with low-light data. Furthermore, our semi-supervised algorithm can be integrated into various detection-based tracking algorithms, expanding the effective dataset size and improving performance compared to training with only limited annotated low-light data.

In detection-based tracking methods, choosing appearance-based or motion-based association methods significantly impacts detection effectiveness, which in turn influences subsequent association performance. Combined with Tab. \ref{tab1}, Tab. \ref{tab2} demonstrates that as detection performance improves, tracking results also improve significantly. Experimental results show that, compared to our baseline OC-SORT\cite{cao2023observation}, the semi-supervised method CRTrack achieves significant performance gains on the DetA, MOTA, HOTA, IDF1, and AssA metrics, demonstrating the effectiveness of CRTrack. Fig. \ref{fig_7} illustrates some example results on the LLMOT validation set. First, the transformation from the original MOT17 data to enhanced low-light data is shown. Then, the ground truth, OC-SORT, and CRTrack results are displayed. In low-light scenarios, CRTrack generates fewer tracking errors than OC-SORT.

Additionally, OC-SORT\cite{cao2023observation}, which heavily relies on detection results to update trajectories, shows suboptimal tracking performance when detection quality is poor, particularly when transferring models trained in normal lighting to low-light environments. Semi-supervised training significantly mitigates this limitation by improving detection robustness. While ByteTrack\cite{zhang2022bytetrack} outperforms CRTrack on DetA and MOTA due to its staged matching strategy, which efficiently leverages low-confidence detection boxes to enhance recall, CRTrack demonstrates superior performance on HOTA and IDF1. By prioritizing trajectory accuracy and robustness through semi-supervised learning, CRTrack adapts more effectively to low-light scenarios, generating higher-quality trajectories with improved stability and association precision.

\begin{figure*}[!t]
\centering
\includegraphics[width=5.3in]{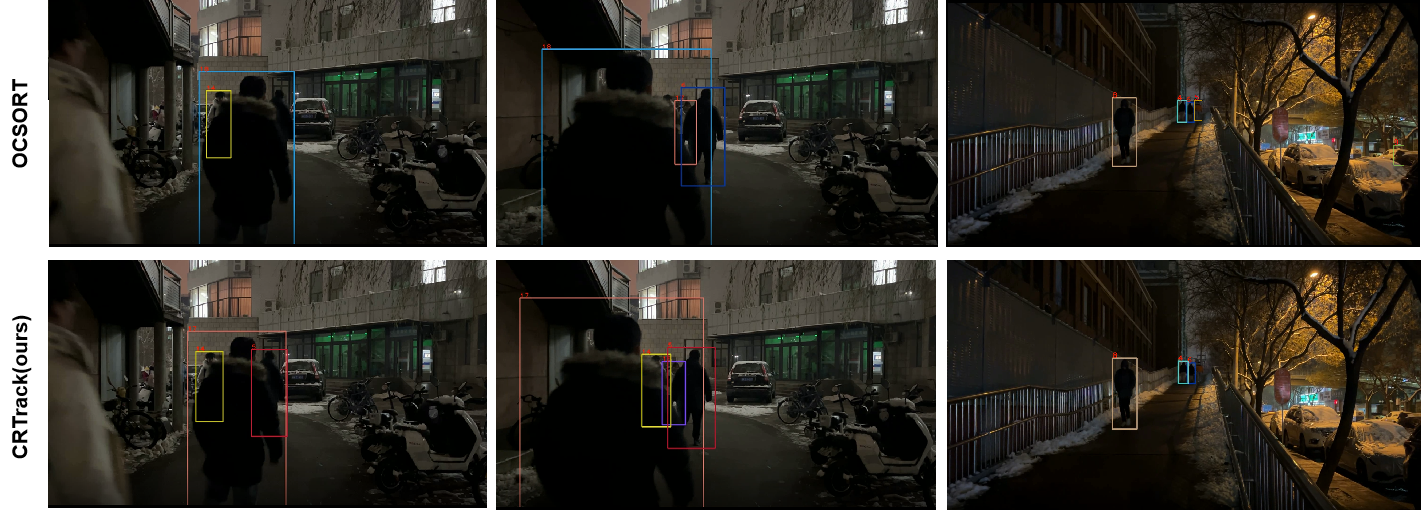}%
\hfil
\caption{Tracking performance of OC-SORT and CRTrack in real low-light scenarios. Our CRTrack generates fewer tracking errors than OC-SORT\cite{cao2023observation}.}
\label{fig_15}
\end{figure*}

In real low-light scenarios, we conducted a comparison test between OC-SORT and CRTrack. As shown in Fig. \ref{fig_15}, OC-SORT exhibits issues such as target loss and less accurate tracking. In contrast, CRTrack demonstrates higher robustness and accuracy under the same conditions, maintaining stable tracking of targets and effectively identifying them even in complex backgrounds and low-light conditions.

\begin{table}
\begin{center}
\small
\caption{Ablation on ASA Weights.}
\label{tab4}
\begin{tabular}{c c c c c c}
\toprule
ASA & DetA & MOTA & HOTA & IDF1 & AssA \\
\midrule
0 & 55.729 & 65.923 & 58.428 & 70.29 & 61.585 \\
1 & 55.747 & 65.942 & 58.807 & \textbf{71.287} & \textbf{62.348} \\ 
2 & \textbf{56.557} & \textbf{66.081} & \textbf{59.011} & 70.881 & 61.993 \\
3 & 55.157 & 64.622 & 57.613 & 68.763 & 60.536 \\
\bottomrule
\end{tabular}
\end{center}
\end{table}

We conducted ablation experiments to validate the effectiveness of our proposed improvements. The results are presented in Tab. \ref{tab3}. In the table, "sem." denotes the use of semi-supervised learning, "ASA" represents consistent adaptive sampling assignment, and "ANU" indicates adaptive network update. "App." refers to the use of appearance features, while "SCD" represents the use of split cosine distance. All improvements are shown to contribute effectively to enhanced performance. As observed in Tab. \ref{tab3}, directly applying semi-supervised training yields suboptimal detection results. Introducing ASA significantly improves the detection evaluation metric DetA. Incorporating ANU further optimizes the model's output, enhancing overall performance. Considering appearance features during association and employing split cosine distance for similarity calculation between tracking and observation features substantially improve IDF1, reduce ID switches, and enhance overall tracking performance.

Furthermore, we conducted experiments to analyze the weight factor $\lambda_{dis}$ for ASA, as shown in Tab. \ref{tab4}. Using OC-SORT\cite{cao2023observation} as the association method, we observed the best results when $\lambda_{dis}$ was set between 1 and 2. Due to random data augmentation applied to the student network during semi-supervised learning, the training results exhibited some variability. Ultimately, we selected $\lambda_{dis}=2$ as the optimal value.
\section{Conclusion}
In this work, we propose the first pedestrian tracking dataset specifically designed for low-light environments. The dataset includes annotated data generated through enhancement techniques and unannotated low-light data collected from real-world scenes. This dataset serves as a critical resource for advancing research in pedestrian tracking under low-light conditions, enabling more comprehensive evaluation and development of tracking algorithms in challenging environments.

We further investigate the task of multi-object tracking in dark scenes. To address the challenges of limited annotated data and the difficulties of manual annotation, we propose a semi-supervised multi-object tracking method. Based on the principle of consistency regularization, we design a teacher-student network framework to enable the model to learn robust features from both labeled and unlabeled data, thereby improving its tracking performance in low-light conditions. Additionally, we introduce a consistent adaptive sampling assignment (ASA) module to mitigate noise effects and enhance consistency between the teacher and student networks. Furthermore, an adaptive update (ANU) module is developed, allowing the student network to update the teacher network more flexibly.

Our work addresses the challenges posed by limited annotated data while enhancing tracking capabilities in low-light environments. By leveraging semi-supervised learning, the proposed method improves feature robustness and tracking accuracy, providing a valuable framework for reliable multi-object tracking in challenging low-light conditions.

\section*{Acknowledgments}
This work was supported by the National Natural Science Foundation of China (61976017 and 61601021), the Beijing Natural Science Foundation (4202056) and the Fundamental Research Funds for the Central Universities (2022JBMC013). The support and resources from the Center for High Performance Computing at Beijing Jiaotong University (http://hpc.bjtu.edu.cn) are gratefully acknowledged.

\bibliographystyle{elsarticle-num} 
\bibliography{ref}
\end{document}